%% file: manusccript.tex
\documentclass[journal,final,letterpaper,twoside,twocolumn]{IEEEtran}
\usepackage{amsmath}
\usepackage{amsfonts}
\usepackage{amssymb}

\usepackage{array}
\usepackage{url}

\usepackage{amsfonts}
\usepackage{mathtools}
\usepackage{dsfont}

\DeclareMathOperator{\Tr}{Tr}

\usepackage{amsmath,graphicx}
\usepackage{dsfont}
\usepackage{amsmath}
\usepackage{amssymb}
\usepackage{amsfonts}
\usepackage{graphicx}
\usepackage{amsmath}
\usepackage[all,poly]{xy}
\usepackage{multirow}
\usepackage{color}
\usepackage[noadjust]{cite}
\usepackage{subfigure}
\usepackage{mathrsfs}
\usepackage{algorithmic}
\usepackage[linesnumbered,lined,ruled]{algorithm2e}
\usepackage{cancel}
\usepackage{url}
\usepackage{lipsum}
\usepackage{stfloats}
\usepackage{epsfig}
\usepackage{bbm}

\newcommand{\bs}{\boldsymbol}

\newcommand{\NoiCovMat}{\bs{\Lambda}}

\input notations.tex

\graphicspath{{figures/}}

\usepackage{tikz}
\usetikzlibrary{arrows}
\tikzset{
    >=stealth',
    image/.style={
           rectangle,
		   fill=yellow!10,
           draw=black, very thick,
           text width=10em,
           minimum height=2em,
           text centered},
    input/.style={
           rectangle,
           rounded corners,
		   fill=yellow!10,
           draw=black, very thick,
           text width=8em,
           minimum height=2em,
           text centered},
    process/.style={
           rectangle,
           rounded corners,
		   fill=blue!10,
           draw=black, very thick,
           text width=8em,
           minimum height=2em,
           text centered},
    object/.style={
           circle,
		   fill=yellow!10,
           draw=black, very thick,
           text width=2em,
           minimum height=1.5em,
           text centered},
    pil/.style={
           ->,
           thick,
           shorten <=2pt,
           shorten >=2pt,}
}

\begin{document}

\title{Hyperspectral pansharpening: a review}

\author{Laetitia~Loncan, Luis B. Almeida, Jos\'e M. Bioucas-Dias, Xavier Briottet, Jocelyn Chanussot, Nicolas Dobigeon, Sophie Fabre, Wenzhi Liao, Giorgio A. Licciardi, Miguel Sim\~oes, Jean-Yves Tourneret, Miguel A. Veganzones, Gemine Vivone, Qi Wei and Naoto Yokoya
\thanks{L.~Loncan is with  Gipsa-lab, Grenoble, France and ONERA (email:laetitia.loncan@onera.fr).}
\thanks{Luis B.~Almeida and J. M. Bioucas-Dias are with  Instituto de Telecomunica\c{c}\~oes, Instituto Superior T\'ecnico, Universidade de Lisboa, (email:\{luis.almeida,bioucas@lx.it.pt\}).}
\thanks{J.~Chanussot, G.~Licciardi, and M.~Veganzones are with Gipsa-lab, Grenoble, France (email:\{Jocelyn.chanussot,Giorgio-Antonino.Licciardi, miguel-angel.veganzones\}@gipsa-lab.grenoble-inp.fr).}
\thanks{X.~Briottet and S.~Fabre are with ONERA, Toulouse, France (email: \{Xavier.briottet, Sophie.fabre\}@onera.fr).}
\thanks{N.~Dobigeon, J.-Y.~Tourneret, and  Q.~Wei are with University of Toulouse, IRIT/INP-ENSEEIHT (email:\{nicolas.dobigeon, jean-yves.tourneret,qi.wei\}@enseeiht.fr).}
\thanks{W.~Liao is with Ghent University, Ghent, Belgium (email:wliao@telin.ugent.be).}
\thanks{M.~Sim\~oes is with Instituto de Telecomunica\c{c}\~oes, Instituto Superior T\'ecnico, Universidade de Lisboa and Gipsa-lab, Grenoble, France (email:miguel.simoes@lx.it.pt).}
\thanks{G.~Vivone is with North Atlantic Treaty Organization (NATO) Science and Technology Organization (STO) Centre for Maritime Research and Experimentation (CMRE) (email:gvivone@unisa.it).}
\thanks{N.~Yokoya is with University of Tokyo (email:yokoya@sal.rcast.u-tokyo.ac.jp).
}}

\maketitle

\begin{abstract}
Pansharpening aims at fusing a panchromatic image with a multispectral one, to generate an image with the high spatial resolution of the former and the high spectral resolution of the latter. In the last decade, many algorithms have been presented in the literature for pansharpening using multispectral data. With the increasing availability of hyperspectral systems, these methods are now being adapted to hyperspectral images. In this work, we compare new pansharpening techniques designed for hyperspectral data with some of the state of the art methods for multispectral pansharpening, which have been adapted for hyperspectral data.  Eleven methods from different classes (component substitution, multiresolution analysis, hybrid, Bayesian and matrix factorization) are analyzed. These methods are applied to three datasets and their effectiveness and robustness are evaluated with widely used performance indicators. In addition, all the pansharpening techniques considered in this paper have been implemented in a MATLAB toolbox that is made available to the community.
\end{abstract}


\section{Introduction}

\IEEEPARstart{I}{n} the design of optical remote sensing systems, owing to the limited amount of incident energy, there are critical tradeoffs between the spatial resolution, the spectral resolution, and signal-to-noise ratio (SNR). For this reason, optical systems can provide data with a high spatial resolution but with a small number of spectral bands (for example, panchromatic data with decimetric spatial resolution or multispectral data with three to four bands and metric spatial resolution, like PLEIADES \cite{PLEIADES}) or with a high spectral resolution but with reduced spatial resolution (for example, hyperspectral data, subsequently referred to as HS data, with more than one hundred of bands and decametric spatial resolution like HYPXIM \cite{HYPXIM}). To enhance the spatial resolution of multispectral data, several methods have been proposed in the literature under the name of pansharpening, which is a form of superresolution. Fundamentally, these methods solve an inverse problem which consists of obtaining an enhanced image with both high spatial and high spectral resolutions from a panchromatic image and a multispectral image. The huge interest of the community on this topic is evidenced by the existence of sessions dedicated to this topic in the most important remote sensing and Earth observation conferences as well as by the launch of public contests, of which the one sponsored by the data fusion committee of the IEEE Geoscience and Remote Sensing society \cite{AlparoneContest2006} is an example.

A taxonomy of pansharpening methods can be found in the literature \cite{Vivo14Tool}, \cite{Aiazzi2012}, \cite{Thom08}. They can be broadly divided into four classes: component substitution (CS), multiresolution analysis (MRA), Bayesian, and variational. The CS approach relies on the substitution of a component (obtained, e.g.,  by a spectral transformation of the data) of the multispectral (subsequently denoted as MS) image by the panchromatic (subsequently denoted as PAN) image. The CS class contains algorithms such as intensity-hue-saturation (IHS) \cite{IHS}, \cite{Tu__01}, \cite{Chav91}, principal component analysis (PCA) \cite{PCA}, \cite{Shet92}, \cite{PCAbis} and Gram-Schmidt (GS) spectral sharpening \cite{Labe00}. The MRA approach is based on the injection of spatial details, which are obtained through a multiscale decomposition of the PAN image into the MS data. The spatial details can be extracted according to several modalities of MRA: decimated wavelet transform (DWT)  \cite{Mallat89}, undecimated wavelet transform (UDWT) \cite{Nason95}, "\`a-trous" wavelet transform (ATWT) \cite{ATWT}, Laplacian pyramid \cite{Burt83}, nonseparable transforms, either based on wavelets (e.g., contourlets \cite{Do05}) or not (e.g., curvelets \cite{Starck07}). Hybrid methods have been also proposed, which use both component substitution and multiscale decomposition, such as guided filter PCA (GFPCA), described in Section \ref{subsec:hybrid}. The Bayesian approach relies on the used of posterior distribution of the full resolution target image given the observed MS and PAN images. This posterior, which is the Bayesian inference engine, has two factors: a) the likelihood function, which is the probability density of the observed MS and PAN images given the target image, and b) the prior probability density of the target image, which promotes target images with desired properties, such as being segmentally smooth. The selection of a suitable prior allows us to cope with the usual ill-posedness  of the pansharpening inverse problems. The variational class is interpretable as particular case of the Bayesian one, where the target image is estimated by maximizing the posterior probability density of the full resolution image. The works \cite{Ballester06}, \cite{Palsson12}, \cite{He2014} are representative of the Bayesian and variational classes. As indicated in Table \ref{schema1}, the CS, MRA, and Hybrid classes of methods are detailed in Sections \ref{subsec:CS}, \ref{subsec:MRA}, and \ref{subsec:hybrid}, respectively. Herein, the Bayesian class is not addressed in the MS+PAN context. It is addressed in detail, however, in Section \ref{subsec:Bayesian} in the context of HS+PAN fusion.\\
\indent With the increasing availability of HS systems, the pansharpening methods are now extending to the fusion of HS and panchromatic images \cite{MoellerVariational}, \cite{Garzelli}, \cite{Alparone15}, \cite{Vivo14HS}. Pansharpening of HS images is still an open issue, and very few methods are presented in the literature to address it. The main advantage of HS image with respect to MS one is the more accurate spectral information they provide, which clearly benefits many applications such as unmixing \cite{Bioucas12}, change detection \cite{changeDetection}, object recognition \cite{ObjetRecognition}, scene interpretation \cite{SceneInterpretation} and classification \cite{ClassifMap}. Several of the methods designed for HS pansharpening were originally designed for the fusion of MS and HS data\cite{YokoyaTGRS2012,Wei2015jstsp,Wei2014icassp,Simoes2014b,Wei2015tgrs}, the MS data constituting the high spatial resolution image. In this case, HS pansharpening can be seen as a particular case, where the MS image is composed of a single band, and thus reduces to a PAN image. In this paper, we divide these methods into two classes: Bayesian methods and matrix factorization based methods. In Section \ref{subsec:Bayesian}, we briefly present the algorithms of \cite{Wei2015jstsp}, \cite{Wei2015tgrs}, and \cite{Simoes2014b} of the former class and in Section \ref{subsec:matrix} the algorithm of \cite{YokoyaTGRS2012} of the latter class.\\
\indent As one may expect, performing pansharpening with HS data is more complex than performing it with MS data. Whereas PAN and MS data are usually acquired almost in the same spectral range, the spectral range of an HS image normally is much wider than the one of the corresponding PAN image. Usually, the PAN spectral range is close to the visible spectral range of  $0.4-0.8\mu$m (for example, the advanced land imager--ALI--instrument acquires PAN data in the range $0.48-0.69\mu$m). The HS range often covers the visible to the shortwave infrared (SWIR) range (for example, Hyperion acquires HS data in the range  $0.4-2.5\mu$m, the range  $0.8-2.5\mu$m being not covered by the PAN data). The difficulty that arises, consists in defining a fusion model that yields good results in the part of the HS spectral range that is not covered by PAN data,in which the high resolution spatial information is missing. This difficulty already existed, to some extent, in MS+PAN pansharpening, but it is much more severe in the HS+PAN case.\\
\indent To the best of the authors' knowledge, there is currently no study comparing different fusion methods for HS data, particularly on datasets where the spectral domain of the HS image is larger than the one of the PAN image. This work aims at addressing this specific issue. The remainder of the paper is organized as follows. Section II reviews the methods under study, i.e., CS, MRA, hybrid, Bayesian, and matrix decomposition approaches. Section III summarizes the quality assessment measures that will be used to assess the image fusion results. Experimental results are presented in Section IV. Conclusions are drawn in Section V.

\section{Hyperspectral Pansharpening Techniques}
\label{sec:HS_pan:tech}
This section presents some of the most relevant methods for HS pansharpening. First, we focus on the adaptation of the popular CS and MRA MS pansharpening methods for HS pansharpening. Later, we consider more recent methods based on Bayesian and matrix factorization approaches. A toolbox containing MATLAB implementations of these algorithms can be found online\footnote{\url{http://openremotesensing.net}}.

Before presenting the different methods, we introduce notation used along the paper. Bold-face capital letters refer to matrices and bold-face lower-case letters refer to vectors. The notation ${\bf X}^k$ refers to the $k$th row of $\bf X$. The operator $()^T$ denotes the transposition operation. Images are represented by matrices, in which each row corresponds to a spectral band, containing all the pixels of that band arranged in lexicographic order. We use the following specific matrices:
\begin{itemize}

\item $\MATima = \left[\bsx_1,\ldots,\bsx_{n}\right] \in \mathbb{R}^{m_\lambda \times n}$ represents
	the full resolution target image with $m_\lambda$ bands and $n$ pixels; $\widehat{{\bf X}}$ represents an estimate of that image.

\item ${\bfY}_{\mathrm{H}} \in \mathbb{R}^{\nbbandima \times m}$, ${\bfY}_{\mathrm{M}}\in\mathbb{R}^{n_{\lambda} \times n}$, and ${\bf P}\in\mathbb{R}^{1\times n}$ represents, respectively, the observed HS, MS, and PAN images, $n_\lambda$ denoting the number of bands of the MS image and $m$ the total number of pixel in the ${\bfY}_{\mathrm{H}}$ image.

\item $\widetilde{{\bfY}}_{\mathrm{H}} \in \mathbb{R}^{m_\lambda \times n}$  represents the HS image ${\bfY}_{\mathrm{H}}$  interpolated at the scale of the PAN image.

\end{itemize}

We denote by $d=\sqrt{m/n}$ the down-sampling factor, assumed to be the same in both spatial dimensions.

\newcommand{\tablecolumnwidth}{0.45\columnwidth}
\begin{table}
\caption{Summary of the different classes of methods considered in this paper. Within parentheses, we indicate the acronym of each method, followed by the number of the section in which that method is described.}
\setlength{\tabcolsep}{0.3mm}
\begin{center}
 \begin{tabular}{|cc|}
  \hline
  \multicolumn{2}{|c|}{\textbf{\textsc{Methods originally designed for MS Pansharpening}}}\\
  \fbox{\noindent
        \begin{tabular}{m{\tablecolumnwidth}}
            \textbf{Component substitution} (CS, \ref{subsec:CS})\\
            Principal Component Analysis  (PCA, \ref{subsubsec:PCA})\\
            Gram Schmidt (GS, \ref{subsubsec:GS})
        \end{tabular}
  }
    &
  \fbox{\noindent
        \begin{tabular}{m{\tablecolumnwidth}}
            \textbf{Multiresolution analysis} (MRA, \ref{subsec:MRA})\\
            Smoothing filter-based intensity modulation (SFIM, \ref{subsubsec:SFIM})\\
            Laplacian pyramid (\ref{subsubsec:LP})
        \end{tabular}
  }
  \\[0.8cm]
  \fbox{\noindent
        \begin{tabular}{m{\tablecolumnwidth}}
            \textbf{Hybrid methods} (\ref{subsec:hybrid})\\
            Guided Filter PCA (GFPCA)
        \end{tabular}
  }
    &
  \fbox{\noindent
        \begin{tabular}{m{\tablecolumnwidth}}
            \textbf{Bayesian methods}\\
             \emph{Not discussed in this paper}
        \end{tabular}
  } \\[0.4cm]
\end{tabular}
\begin{tabular}{|cc|}
  \hline
  \hline
  \multicolumn{2}{|c|}{\textbf{\textsc{Methods originally designed for HS Pansharpening}}}\\
  \fbox{\noindent
        \begin{tabular}{m{\tablecolumnwidth}}
            \textbf{Bayesian Methods} (\ref{subsec:Bayesian})\\
            Naive Gaussian prior  (\ref{subsubsec:Naive_Gaussian})\\
            Sparsity promoting prior (\ref{subsubsec:Sparsity_Gaussian})\\
            HySure (\ref{subsubsec:Hysure})
        \end{tabular}
  }
    &
  \fbox{\noindent
        \begin{tabular}{m{\tablecolumnwidth}}
            \textbf{Matrix Factorization} (\ref{subsec:matrix})\\
            Coupled Non-negative Matrix Factorization (CNMF)\\
            \quad\\
        \end{tabular}
  } \\[0.7cm]
  \hline
\end{tabular}
\end{center}
\label{schema1}
\end{table}


\subsection{Component Substitution}
\label{subsec:CS}

CS approaches rely upon the projection of the higher spectral resolution image into another space, in order to separate spatial and spectral information~\cite{Thom08}. Subsequently, the transformed data are sharpened by substituting the component that contains the spatial information with the PAN image (or part of it). The greater the correlation between the PAN image and the replaced component, the less spectral distortion will be introduced by the fusion approach~\cite{Thom08}. As a consequence, a histogram-matching procedure is often performed before replacing the PAN image. Finally, the CS-based fusion process is completed by applying the inverse spectral transformation to obtain the fused image.

The main advantages of the CS-based fusion techniques are the following: $i)$ high fidelity in rendering the spatial details in the final image~\cite{Aiaz07}, $ii)$ fast and easy implementation~\cite{Tu__01}, and $iii)$ robustness to misregistration errors and aliasing~\cite{Baronti2011}. On the negative side, the main shortcoming of this class of techniques is the generation of a significant spectral distortion, cause by the spectral mismatch between the PAN and the HS spectral ranges~\cite{Thom08}.

Following~\cite{Vivo15,Vivo14Tool}, a  formulation of the CS fusion scheme is given by
\begin{align}
\widehat{{\bf X}}^{k}& =\widetilde{{\bf Y}}_{\mathrm{H}}^{k}+g_{k}\left(\mathbf{P}-\mathbf{O}_{L}\right),
\label{eq:genaralInjectionModel}%
\end{align}
 for $k=1,\dots,m_\lambda$, where $\widehat{{\bf X}}^{k}$ denotes the $k$th band of the estimated full resolution target image,
 $\mathbf{g}=[g_{1},{\dots},g_{m_\lambda}]^T$ is a vector containing the \emph{injection gains}, and
 $\mathbf{O}_{L}$ is defined as
\begin{equation}
\mathbf{O}_{L}=\overset{m_\lambda}{\underset{i=1}{\sum}}w_{i}\widetilde{{\bf Y}}_{\mathrm{H}}^{i},
\label{eq:linearPLP}%
\end{equation}
where the weights $\mathbf{w}=[w_{1},\dots,w_{i},\dots,w_{m_\lambda}]^T$ measure the spectral overlap among the spectral bands and the PAN image~\cite{Tu__04,Thom08}.

The CS family includes many popular pansharpening approaches. In \cite{Vivo14HS}, three approaches based on \emph{principal component analysis} (PCA)~\cite{Chav91} and \emph{Gram-Schmidt}~\cite{Labe00,Aiaz07} transformations have been compared for sharpening HS data. A brief description of these techniques follows.

\subsubsection{Principal Component Analysis}
\label{subsubsec:PCA}
\emph{PCA} is a spectral transformation widely employed for pansharpening applications~\cite{Chav91}. It is achieved through a rotation of the original data (i.e., a linear transformation) that yields the so-called principal components (PCs). The hypothesis underlying its application to pansharpening is that the spatial information (shared by all the channels) is concentrated in the first PC, while the spectral information (specific to each single band) is accounted for the other PCs. The whole fusion process can be described by the general formulation stated by Eqs. \eqref{eq:genaralInjectionModel} and \eqref{eq:linearPLP}, where the vectors $\mathbf{w}$ and $\mathbf{g}$ of coefficient vectors are derived by the PCA procedure applied to the HS image.
\subsubsection{Gram-Schmidt}
\label{subsubsec:GS}
The \emph{Gram-Schmidt} transformation, often exploited in pansharpening approaches, was initially proposed in a patent by Kodak~\cite{Labe00}. The fusion process starts by using, as the component, a synthetic low resolution PAN image $\mathbf{I}_{L}$ at the same spatial resolution as the HS image\footnote{GS is a more general method than PCA. PCA can be obtained, in GS, by using the first PC as the low resolution panchromatic image~\cite{Aiaz09}.}. A complete orthogonal decomposition is then performed, starting with that component. The pansharpening procedure is completed by substituting that component with the PAN image, and inverting the decomposition. This process is expressed by (\ref{eq:genaralInjectionModel}) using the gains~\cite{Aiaz07}
\begin{equation}
g_{k}=\frac{\textrm{cov}(\widetilde{{\bf Y}}_{\mathrm{H}}^{k},\mathbf{O}_{L})}{\textrm{var}(\mathbf{O}_{L})}
\end{equation}
for $k=1,\dots,m_\lambda$, where $\text{cov}\left(\cdot,\cdot\right)$ and $\text{var}\left(\cdot\right)$ denote the covariance and variance operations. Different algorithms are obtained by changing the definition of the weights in (\ref{eq:linearPLP}). The simplest way to obtain this low-resolution PAN image simply consists of averaging the HS bands (i.e., by setting $w_{i}=1/m_\lambda$, for $i=1,\dots,m_\lambda$). In~\cite{Aiaz07}, the authors proposed an enhanced version, called \emph{GS Adaptive} (\emph{GSA}), in which $\mathbf{I}_{L}$ is generated by the linear model in~(\ref{eq:linearPLP}) with weights estimated by the minimization of the mean square error between the estimated component and a filtered and downsampled version of the PAN image.

\subsection{Multiresolution Analysis}
 \label{subsec:MRA}
 Pansharpening methods based on MRA apply a spatial filter to the PAN image for generating details to be injected into the HS data. The main advantages of the MRA-based fusion techniques are the following: $i)$ temporal coherence \cite{Aiazzi2012} (see Sect.27.4.4), $ii)$ spectral consistency, and $iii)$ robustness to aliasing, under proper conditions \cite{Baronti2011}. On the negative side, the main shortcomings are $i)$ the implementation is more complicated due to the design of spatial filters, $ii)$ the computational burden is usually larger when compared to CS approaches. The fusion step is summarized as ~\cite{Vivo15,Vivo14Tool}
\begin{align}
\widehat{{\bf X}}^{k}& =\widetilde{{\bf Y}}_{\mathrm{H}}^{k}+\mathbf{G}_k \otimes\left(\mathbf{P}-\mathbf{P}_{L}\right),
\label{eq:MRA}
\end{align}
for $k=1,\dots,m_\lambda$, where $\mathbf{P}_{L}$ denotes a low-pass version of $\bf P$, and the symbol $\otimes$ denotes element-wise multiplication.
Furthermore, an equalization between the PAN image and the HS spectral bands is often required. $\mathbf{P}-\mathbf{P}_{L}$ is often called the \emph{details} image, because it is a high-pass version of $\mathbf{P}$, and Eq.~\eqref{eq:MRA} can be seen as describing the way to \emph{inject} details into each of the bands of the HS image. According to (\ref{eq:MRA}), the approaches belonging to this category can differ in $i)$ the type of PAN low pass image $\mathbf{P}_{L}$ that is used, and $ii)$ the definition of the gain coefficients  $\mathbf{G}_{k}$. Two common options for defining the gains are:
\begin{enumerate}
\item $\mathbf{G}_k = \mathbbm{1}$ for $k=1,\dots,m_\lambda$, where $\mathbbm{1}$ is an appropriately sized matrix with all elements equal to 1. This choice identifies the so-called \emph{additive} injection scheme;
\item $\mathbf{G}_k = \widetilde{{\bf Y}}_{\mathrm{H}}^{k}\oslash\mathbf{P}_{L}$ for $k=1,\dots,m_\lambda$,
where the symbol $\oslash$ denotes element-wise division. In this case, the details are weighted by the ratio between the upsampled HS image and the low-pass filtered PAN one, in order to reproduce the local intensity contrast of the PAN image in the fused image~\cite{Vivo14}. This coefficient selection is often referred to as \emph{high pass modulation} (HPM) method or \emph{multiplicative} injection scheme. Some possible numerical issues could appear due to the division between $\widetilde{{\bf Y}}_{\mathrm{H}}^{k}$ and $\mathbf{P}_{L}$ for low value of $\mathbf{P}_{L}$ creating fused pixel with very high value. In our toolbox this problem is addressed by clipping these values by using the information given by the dynamic range.
\end{enumerate}

In the case of HS pansharpening, some further considerations should be taken into account. Indeed, the PAN and HS images are rarely acquired with the same platform. Thus, the ratio between the spatial resolutions of the PAN and HS images may not always be an integer number, or a power of two. This implies that some of the conventional approaches initially developed for MS images cannot be extended in a simple way to HS images (for example, dyadic wavelet-based algorithms cannot be applied in these conditions).

\subsubsection{Smoothing Filter-based Intensity Modulation (SFIM)}
\label{subsubsec:SFIM}
The  direct implementation of Eq.~(\ref{eq:MRA}) consists of applying a single linear time-invariant (LTI) low pass filter (LPF) $h_{LP}$ to the PAN image $\mathbf{P}$ for obtaining $\mathbf{P}_{L}$. Therefore, we have
\begin{equation}
\widehat{{\bf X}}^{k} =\widetilde{{\bf Y}}_{\mathrm{H}}^{k}+g_{k}\left(\mathbf{P}-\mathbf{P} \ast h_{LP}\right)
\label{eq:HPF2}
\end{equation}
for $k=1,\dots,m_\lambda$, where the symbol $\ast$ denotes the convolution operator. The SFIM algorithm~\cite{Liu_00} sets $h_{LP}$ to a simple box (i.e., an averaging) filter and exploits HPM as the details injection scheme.

\subsubsection{Laplacian Pyramid}
\label{subsubsec:LP}
The low-pass filtering needed to obtain the signal $\mathbf{P}_{L}$ at the original HS scale can be performed in more than one step. This is commonly referred to as pyramidal decomposition and dates back to the seminal work of Burt and Adelson~\cite{Burt83}. If a Gaussian filter is used to low-pass filter the images in each step, one obtains a so-called \emph{Gaussian pyramid}. The differences between consecutive levels of a Gaussian pyramid define the so-called \emph{Laplacian pyramid}. The suitability of the latter to the pansharpening problem has been shown in~\cite{Alpa03}. Indeed, Gaussian filters can be tuned to closely match the sensor modulation transfer function (MTF). In this case, the unique parameter that characterizes the whole distribution is the Gaussian's standard deviation, which is determined from sensor-based information (usually from the value of the amplitude response at the Nyquist frequency, provided by the manufacturer). Both \emph{additive} and \emph{multiplicative} details injection schemes have been used in this framework~\cite{Vivo14,Aiaz06}. They will be referred to as \emph{MTF - Generalized Laplacian Pyramid} (MTF-GLP)~\cite{Aiaz06} and \emph{MTF-GLP with High Pass Modulation} (MTF-GLP-HPM)~\cite{Vivo14}, respectively.

\subsection{Hybrid Methods}
\label{subsec:hybrid}
Hybrid approaches use concepts from different classes of methods, namely from CS and MRA ones, as explained next.

 \subsubsection{Guided Filter PCA (GFPCA)}
 \label{subsubsec:GFPCA}
 One of the main challenges for fusing low-resolution HS and high-resolution PAN/RGB data is to find an appropriate balance between spectral and spatial preservation. Recently, the guided filter~\cite{IEEEhowto:He13} has been used in many applications (e.g. edge-aware smoothing and detail enhancement), because of its efficiency and strong ability to transfer the structures of the guidance image to the filtering output. Its application to HS data can be found in~\cite{IEEEhowto:Kang14}, where the guided filter was applied to transfer the structures of the principal components of the HS image to the initial classification maps.

\begin{figure}[b!]
\includegraphics[width=.95\linewidth]{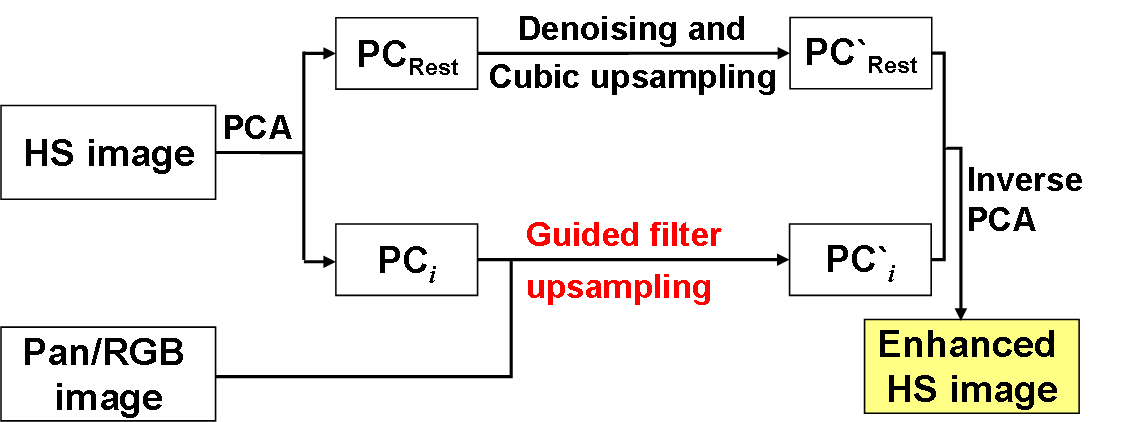}
\caption{Fusion of HS and PAN/RGB images with the GFPCA framework.}
\label{Liaofig1}
\end{figure}

Here, we briefly describe an image fusion framework which uses a guided filter in the PCA domain (GFPCA) \cite{IEEEhowto:Liao15}. The approach won the ``Best Paper Challenge'' award at the 2014 IEEE data fusion contest \cite{IEEEhowto:Liao15}, by fusing a low spatial resolution thermal infrared HS image and a high spatial resolution, visible RGB image associated with the same scene. Fig.~\ref{Liaofig1} shows the framework of GFPCA. Instead of using CS, which may cause spectral distortions, GFPCA uses a high resolution PAN/RGB image to guide the filtering process aimed at obtaining superresolution. In this way, GFPCA does not only preserve the spectral information from the original HS image, but also transfers the spatial structures of the high resolution PAN/RGB image to the enhanced HS image. To speed up the processing, GFPCA first uses PCA to decorrelate the bands of the HS image, and to separate the information content from the noise. The first $p\ll m_\lambda$ PCA channels contain most of the energy (and most of the information) of an HS image, and the remaining $m_\lambda-p$ PCA channels mainly contain noise (recall that $m_\lambda$ is the number of spectral bands of the HS image). When applied to these noisy (and numerous) $m_\lambda-p$ channels, the guided filter amplifies the noise and causes a high computational cost in processing the data, which is undesirable. Therefore, guided filtering is used to enlarge only the first $k$ PCA channels, preserving the structures of the PAN/RGB image, while cubic interpolation is used to upsample the remaining channels.

Let $\textrm{PC}_i$, with ($i\leq p$), denote the  $i$th  PC channel obtained from the HS image ${\bf Y}_{\mathrm{H}}$, with its resolution increased to that of the guided image $\bf Y$ ($\bf Y$ may be a PAN or an RGB image) through bicubic interpolation. The output of the filtering, $\textrm{PC}_i^\prime$, can be represented as an affine transformation of $\bf Y$ in a local window $\omega_j$ of size $(2d+1)\times(2d+1)$ as follows:
\begin{equation}
\textrm{PC}_i^\prime=a_j {\bf Y}+b_j, \;\; \forall i\in \omega_j.
     \label{eq1}
\end{equation}
The above model ensures that the output $\textrm{PC}_i^\prime$ has an edge only if the guided image $\bf Y$ has an edge, since $\nabla (\textrm{PC}_i^\prime)=a\nabla \bf Y$. The following cost function is used to determine the coefficients $a_j$ and $b_j$:
\begin{equation}
E(a_j,b_j)=\sum_{i\in \omega_j} \left[(a_j{\bf Y}+b_j - \textrm{PC}_i)^2+\epsilon a_j^2 \right],
  \label{eq2}
\end{equation}
where $\epsilon$ is a regularization parameter determining the degree of blurring for the guided filter. For more details about the guided filtering scheme, we invite the reader to consult \cite{IEEEhowto:He13}. The cost function $E$ leads the term $a_j {\bf Y}+b_j$ to be as close as possible to $\textrm{PC}_i$, in order to ensure the preservation of the original spectral information. Before applying inverse PCA, GFPCA also removes the noise from the remaining PCA channels $PC_{Rest}$ using a soft-thresholding scheme (similarly to~\cite{IEEEhowto:Liao13}), and increases their spatial resolution to the resolution of the PAN/RGB image using cubic interpolation only (without guided filtering).

\subsection{Bayesian Approaches}
\label{subsec:Bayesian}
The fusion of HS and high spatial resolution images, e.g., MS or PAN images, can be conveniently formulated within the Bayesian inference framework. This formulation allows an intuitive interpretation of the fusion process via the posterior distribution of the Bayesian fusion model. Since the fusion problem {is usually ill-posed, the Bayesian methodology offers a convenient way to regularize the problem by defining an appropriate prior distribution for the scene of interest. Following this strategy, different Bayesian estimators for fusing co-registered high spatial-resolution MS and high spectral-resolution HS images have been designed \cite{Hardie2004,Zhang2009,Joshi2010,Wei2015jstsp,Wei2014icassp,Wei2014ICIP,Wei2015tgrs,
Simoes2014, Simoes2014b}. The observation models associated with the HS and MS images can
be written as follows \cite{Molina1999,Hardie2004,Molina2008}


\begin{equation}
\begin{array}{ll}
\label{eq:HS_MS_obs}
\bfY_{\mathrm{H}} =  \MATima \bf{BS} + \bfN_{\mathrm{H}} \\
\bfY_{\mathrm{M}} =  \bfR \MATima + \bfN_{\mathrm{M}}
\end{array}
\end{equation}
where $\bf X$, $\bfY_{\mathrm{H}}$, and $\bfY_{\mathrm{M}}$ were defined in Section \ref{sec:HS_pan:tech}, and
\begin{itemize}
\item $\bfB \in \mathbb{R}^{n \times n}$ is a cyclic convolution operator, corresponding to the spectral response of the HS sensor expressed in the resolution of the MS or PAN image,
\item ${\bf S}\in\mathbb{R}^{n\times m}$ is a down-sampling matrix with down-sampling factor $d$,
\item ${\bfR}\in\mathbb{R}^{n_{\lambda} \times \nbbandima}$ is the spectral response of the MS or PAN sensor,
\item $\bfN_{\mathrm{H}}$ and $\bfN_{\mathrm{M}}$ are the HS and MS noises, assumed to have zero mean Gaussian distributions with covariance matrices $\NoiCovMat_{\mathrm{H}}$ and $\NoiCovMat_{\mathrm{M}}$, respectively.
\end{itemize}

For the sake of generality, the formulation in this section assumes that the observed data is the pair of matrices $(\bfY_{\mathrm{H}}, \bfY_{\mathrm{M}})$. Since a PAN image can be} represented by $\bfY_{\mathrm{M}}$ with $n_\lambda = 1$, the observation model \eqref{eq:HS_MS_obs} covers the HS+PAN fusion problem considered in this paper.

Using geometrical considerations well grounded in the HS imaging literature devoted to the linear unmixing problem \cite{Bioucas12}, the high spatial resolution HS image to be estimated is assumed to live in a low dimensional subspace. This hypothesis is very reliable when the observed scene is composed of a finite number of macroscopic materials (called \emph{endmembers}). Based on the model \eqref{eq:HS_MS_obs} and on the low dimensional subspace assumptions, the distributions of $\bfY_{\mathrm{H}}$ and $\bfY_{\mathrm{M}}$ can be expressed as follows
\begin{equation}
\begin{array}{ll}
\label{eq:likelihood}
\bfY_{\mathrm{H}}|\bfU \sim \mathcal{MN}_{\nbbandima,m}({\bfH}\bfU \bfB \bfS, \NoiCovMat_{\mathrm{H}}, \Id{m}), \\
\bfY_{\mathrm{M}}|\bfU \sim \mathcal{MN}_{n_{\lambda},n}(\bfR {\bfH}\bfU, \NoiCovMat_{\mathrm{M}}, \Id{n})
\end{array}
\end{equation}
where $\mathcal{MN}$ represents the matrix normal distribution \cite{MNDist}, the target image is $\bf X=HU$, with ${\bf H}\in\mathbb{R}^{m_\lambda\times\widetilde{m}_\lambda}$ containing in its columns a basis of the signal subspace of size $\widetilde{m}_\lambda\ll m_\lambda $ and $\bfU\in \mathbb{R}^{\wtm_{\lambda} \times n}$ contains the representation coefficients of $\bf X$ with respect to $\bf H$.
The subspace transformation matrix $\bfH$ can be obtained via different approaches, e.g., PCA \cite{Farrell2005} or vertex component analysis \cite{Nascimento2005}.

According to Bayes' theorem and using the fact that the noises ${\bf N}_H$ and ${\bf N}_M$ are independent, the posterior distribution of $\bfU$ can be written as
\begin{equation}
\label{eq:posterior_joint}
  p\left(\bfU|\bfY_{\mathrm{H}},\bfY_{\mathrm{M}}\right) \propto p\left(\bfY_{\mathrm{H}}|\bfU\right)p\left(\bfY_{\mathrm{M}}|\bfU\right)p\left(\bfU\right)
\end{equation}
or equivalently\footnote{We use the symbol $\doteq$ to denote equality apart from an additive constant. The additive constants are irrelevant, since the functions under consideration are to be optimized, and the additive constants don't change the locations of the optima.}
\begin{equation}
\label{eq:obj_function}
\begin{array}{ll}
-\log p\left(\bfU|\bfY_{\mathrm{H}},\bfY_{\mathrm{M}}\right)\doteq
          \underbrace{\frac{1}{2}\big\|\NoiCovMat_{\mathrm{H}}^{-\frac{1}{2}}({\bf Y}_{\mathrm{H}}-{\bf HUBS})\big\|_F^2}_{\substack{\text{HS data term}\\ \doteq \log p({\bf Y}_{\mathrm{H}}|{\bfU})}}  +\\
          \underbrace{\frac{1}{2}\big\|\NoiCovMat_{\mathrm{M}}^{-\frac{1}{2}}({\bf Y}_{\mathrm{M}}-{\bf RHU})\big\|_F^2}_{\substack{\text{MS data term} \\ \doteq \log p({\bf Y}_{\mathrm{M}}|{\bfU})}}  +
          \underbrace{\lambda\phi(\bfU)}_{\substack{\text{regularizer}\\ \doteq \log p({\bfU})}}
\end{array}
\end{equation}
where $\| \mathbf{X} \|_F \; {\buildrel\rm def\over=} \; \sqrt{\Tr(\mathbf{X} \mathbf{X}^T)}$ is the Frobenius norm of $\mathbf{X}$. An important quantity in the negative log-posterior \eqref{eq:obj_function} is the penalization term $\phi(\bfU)$ which allows the inverse problem \eqref{eq:HS_MS_obs} to be regularized. The next sections discuss different ways of defining this penalization term.

\subsubsection{Naive Gaussian prior}
\label{subsubsec:Naive_Gaussian}
Denote as  $\bsu_i$ ($i=1,\cdots,n$) the columns of the matrix $\bfU$ that are assumed to be mutually independent and are assigned the following Gaussian prior distributions
\begin{equation}
\label{eq:prior_scene}
  p\left(\bsu_i|\imam{i},\imacovmat{i}\right)=\calN\left(\imam{i},\imacovmat{i}\right)
\end{equation}
where $\imam{i}$ is a fixed image defined by the interpolated HS image projected into the subspace of interest, and $\imacovmat{i}$ is an unknown hyperparameter matrix. Different interpolations can be investigated to build the mean vector $\imam{i}$. In this paper, we have followed the strategy proposed in \cite{Hardie2004}. To reduce the number of parameters to be estimated, the matrices $\imacovmat{i}$ are assumed to be identical, i.e., $\imacovmat{1}=\cdots=\imacovmat{{n}}=\Covsub$. The hyperparameter $\Covsub$ is assigned a proper prior and is estimated jointly with the other parameters of interest.
To infer the parameter of interest, namely the projected highly resolved HS image $\bfU$, from the posterior distribution $p\left(\bfU|\bfY_{\mathrm{H}},\bfY_{\mathrm{M}}\right)$, several algorithms have been proposed. In \cite{Wei2015jstsp,Wei2014icassp}, a Markov chain Monte Carlo (MCMC) method is exploited to generate a collection of $N_{\textrm{MC}}$ samples that are asymptotically distributed according to the target posterior. The corresponding Bayesian estimators can then be approximated using these generated samples. For instance, the minimum mean square error (MMSE) estimator of $\bfU$ can be approximated by an empirical average of the generated samples $\widehat{{\bf U}}_{\text{MMSE}} \approx \frac{1}{N_{\textrm{MC}}-N_{\textrm{bi}}}
\sum_{t=N_{\textrm{bi}}+1}^{N_{\textrm{MC}}} {\bfU}^{(t)}$, where $N_{\textrm{bi}}$ is the number of burn-in iterations required to reach the sampler convergence, and ${\bfU}^{(t)}$ is the image generated in the $t$th iteration. The highly-resolved HS image can finally be computed as $\widehat{{\bf X}}_{\text{MMSE}}= {\bf H}\widehat{{\bf U}}_{\text{MMSE}}$. An extension of the proposed algorithm has been proposed in \cite{Wei2014ICIP} to handle the specific scenario of an unknown sensor spectral response. In \cite{Wei2015whispers}, a deterministic counterpart of this MCMC algorithm has been developed, where the Gibbs sampling strategy of \cite{Wei2015jstsp} has been replaced with a block coordinate descent method to compute the maximum a posteriori (MAP) estimator. Finally, very recently, a Sylvester equation-based explicit solution of the related optimization problem has been derived in \cite{Wei2015sub}, leading to a significant decrease of the computational complexity.

\subsubsection{Sparsity promoted Gaussian prior}
\label{subsubsec:Sparsity_Gaussian}
Instead of incorporating a simple Gaussian prior or smooth regularization for the HS and MS
fusion \cite{Hardie2004,Zhang2009,Wei2014icassp}, a sparse representation can be used to
regularize the fusion problem. More specifically, image patches of the target image (projected onto the subspace defined by $\bfH$) are represented as a sparse linear combination of elements from an appropriately chosen over-complete dictionary with columns referred to as atoms. Learning the dictionary from the observed images instead of using predefined bases \cite{Mallat1999,Starck2002,AHMED1974} generally improves image representation \cite{Elad2006}, which is preferred in most scenarios. Therefore, an adaptive sparse image-dependent regularization can be explored to solve the fusion problem \eqref{eq:HS_MS_obs}. In \cite{Wei2015tgrs}, the following regularization term was introduced:
\begin{equation}
\label{eq:regul}
\phi(\bfU)\propto-\log p\left(\bfU\right)\doteq\frac{1}{2}\sum_{k=1}^{\wtm_{\lambda}} \big\|{\bfU_k -\calP\left(\bar\bfD_k \bar\bfA_k\right) \big\|}_F^2~,
\end{equation}
where
\begin{itemize}

\item $\bfU_k \in \mathbb{R}^{n}$ is the $k$th band (or row) of $\bfU \in \mathbb{R}^{\wtm_{\lambda} \times n}$, with $k=1,\ldots,\wtm_{\lambda}$,

\item $\calP(\cdot): \mathbb{R}^{n_{\textrm{p}} \times n_{\textrm{pat}}} \mapsto \mathbb{R}^{n \times 1}$ is a linear operator that averages
the overlapping patches\footnote{A decomposition into overlapping patches was adopted, to prevent the occurrence of blocking artifacts \cite{Guleryuz2006}.} of each band, $n_\textrm{pat}$ being the number of patches associated with the $i$th band,

\item $\bar\bfD_i \in \mathbb{R}^{n_{\textrm{p}} \times n_{\textrm{at}}}$ is the overcomplete dictionary, whose columns
are basis elements of size $n_{\textrm{p}}$ (corresponding to the size of a patch), $n_{\textrm{at}}$ being the number of dictionary atoms, and

\item $\bar\bfA_i \in \mathbb{R}^{n_{\textrm{at}} \times n_{\textrm{pat}}}$ is the code of the $i$th band.

\end{itemize}

Inspired by hierarchical models frequently encountered in Bayesian inference \cite{Gelman2013}, a second level of hierarchy can be considered in the Bayesian paradigm
by including the code $\bfA$ within the estimation, while fixing the support $\bar{\bs{\Omega}} \triangleq \left\{ \bar{\bs{\Omega}}_1,\ldots, \bar{\bs{\Omega}}_{\wtm_{\lambda}}\right\}$ of the code $\bfA$. Once $\bar\bfD$, $\bar{\bs{\Omega}}$ and $\bfH$ have been learned from the HS and MS data, maximizing the posterior distribution of $\bfU$ and $\bfA$ reduces to a standard constrained quadratic optimization problem with respect to (w.r.t.) $\bfU$ and $\bfA$. The resulting optimization problem is difficult to solve due to its large dimension and due to the fact that the linear operators ${\bf H(\cdot)BD}$ and $\calP(\cdot)$ cannot be easily diagonalized. To cope with this difficulty, an optimization technique that alternates minimization w.r.t. $\bfU$ and $\bfA$ has been introduced in \cite{Wei2015tgrs} (where details on the learning of $\bar\bfD$, $\bar{\bs{\Omega}}$ and $\bfH$ can be found). In \cite{Wei2015sub}, the authors show that the minimization w.r.t. $\bfU$ can be achieved analytically, which greatly accelerates the fusion process.

\subsubsection{HySure}
\label{subsubsec:Hysure}
The works \cite{Simoes2014, Simoes2014b} introduce a convex regularization problem which can be seen under a Bayesian framework. The proposed method uses a form of vector total variation (VTV)~\cite{Bresson2008} for the regularizer $\phi(\bfU)$, taking into account both the spatial and the spectral characteristics of the data. In addition,  another convex problem is formulated to estimate the relative spatial and spectral responses of the sensors $\bfB$ and $\bfR$ from the data themselves. Therefore, the complete methodology can be classified as a blind superresolution method, which, in contrast to the  classical blind linear inverse problems, is tackled by solving two convex problems.

The VTV regularizer  (see \cite{Bresson2008}) is given by
\begin{equation}
\phi\big(\mathbf{U} \big) \; = \; \sum_{j=1}^{n} \sqrt{\sum_{k=1}^{\widetilde{m}_\lambda} \Big\{\big[(\mathbf{U} \mathbf{D}_h)_{j}^k\big]^2 + \big[(\mathbf{U} \mathbf{D}_v)_{j}^k\big]^2\Big\}},
\end{equation}
where $\mathbf{A}^k_j$ denotes the element in the $k$th row and $j$th column of matrix $\mathbf{A}$, and the products by matrices $\mathbf{D}_h$ and $\mathbf{D}_v$ compute the horizontal and vertical discrete differences of an image, respectively, with periodic boundary conditions.

The HS pansharpened image is the solution to the following optimization problem

\begin{align} \label{eq:optimizationproblem}
& \underset{\mathbf{U}}{\text{minimize}}
& & \frac{1}{2}\Big\|\mathbf{Y}_{\mathrm{H}} - \mathbf{H}\mathbf{U}\mathbf{B}\mathbf{S} \Big\|_F^2 + \frac{\lambda_{m}}{2}\Big\|\bfY_{\mathrm{M}} - \mathbf{R}\mathbf{H}\mathbf{U} \Big\|_F^2 \nonumber \\
& & & \quad + \lambda_{\phi} \phi \big(\mathbf{U}),
\end{align}
where $\lambda_m$ and $\lambda_{\phi}$ control the relative weights of the different terms. The optimization problem
\eqref{eq:optimizationproblem} is hard to solve, essentially for three reasons: the downsampling operator $\mathbf{BS}$ is not diagonalizable, the regularizer $\phi \big(\mathbf{U})$ is nonquadratic and nonsmooth, and the target image has a very large size. These difficulties were tackled by solving the problem via the split augmented lagrangian shrinkage algorithm (SALSA)~\cite{Afonso2011}, an instance of ADMM. As an alternative, the main step of the ADMM scheme can be conducted using an explicit solution of the corresponding minimization problem, following the strategy in \cite{Wei2015sub}.

The relative spatial and spectral responses $\mathbf{B}$ and $\mathbf{R}$ were estimated by solving the following optimization problem:
\begin{equation}
\label{eq:B_R_optim}
\begin{aligned}
\underset{\mathbf{B,R}}{\text{minimize}} &&
\big\|\mathbf{R} \mathbf{Y}_{\mathrm{H}} - \mathbf{Y}_{\mathrm{M}} \mathbf{B} \mathbf{S} \big\|^2 + \lambda_B \phi_B(\mathbf{B}) + \lambda_R \phi_R(\mathbf{R})
\end{aligned}
\end{equation}
where $\phi_B(\cdot)$ and $\phi_r(\cdot)$ are quadratic regularizers and $\lambda_b,\lambda_R\geq 0$ are their respective regularization parameters.

\subsection{Matrix factorization}
\label{subsec:matrix}

The matrix factorization approach for HS+MS fusion essentially exploits two facts: 1) A basis or dictionary $\bf H$ for the signal subspace can be learned from the HS observed image $\bfY_{\mathrm{H}}$, yielding the factorization $\bf X=HU$; 2) using this decomposition in the second equation of \eqref{eq:likelihood} and for negligible noise, i.e., ${\bf N}_ {\text{M}}\simeq \bf 0$, we have $\bfY_{\mathrm{H}}={\bf RH U}$. Assuming that the columns of $\bf RH$ are full rank or that the columns of $\bf U$ admit a sparse representation w.r.t. the columns of $\bf RH$, then we can recover the true solution, denoted by $\widehat{\bf U}$, and use it to compute the target image as $\widehat{\bf X} = {\bf H}\widehat{\bf U}$. The works \cite{Berne2010,Kawakami2011,Charles2011,YokoyaTGRS2012, Huang2014, miguelVeng2014} are representative of this line of attack. In what follow, we detail the application of the coupled nonnegative matrix factorization (CNMF) \cite{YokoyaTGRS2012} to the HS+PAN fusion problem.

The CNMF was proposed for the fusion of low spatial resolution HS and high spatial resolution MS data to produce fused data with high spatial and spectral resolutions \cite{YokoyaTGRS2012}. It is applicable to HS pansharpening as a special case, in which the higher spatial resolution image has a single band \cite{YokoyaWHISPERS2011}. CNMF alternately unmixes both sources of data to obtain the endmember spectra and the high spatial resolution abundance maps.

To describe this method, it is convenient to first briefly introduce linear mixture models for HS images. These models are commonly used for spectral unmixing, owing to their physical effectiveness and mathematical simplicity \cite{Bioucas12}. The spectrum at each pixel is assumed to be a linear combination of several endmember spectra. Therefore, $\mathbf{X}\in \mathbb{R}^{m_\lambda \times n}$ is formulated as

\begin{equation}
\label{eq:LSMM}
\mathbf{X} = \mathbf{HU}
\end{equation}

where $\mathbf{H} \in \mathbb{R}^{m_\lambda \times p}$ is the signature matrix, containing the spectral representations of the endmembers, and $\mathbf{U} \in \mathbb{R}^{p \times n}$ is the abundance matrix, containing the relative abundances of the different endmembers at the various pixels, with $p$ representing the number of endmembers. By substituting \eqref{eq:LSMM} into \eqref{eq:HS_MS_obs}, $\mathbf{Y}_{\mathrm{H}}$ and $\mathbf{Y}_{\mathrm{M}}$ can be approximated as

\begin{equation}
\begin{array}{ll}
\label{eq:approx_LSMM}
\mathbf{Y}_{\mathrm{H}} \approx \mathbf{HU}_{\mathrm{H}} \\
\mathbf{Y}_{\mathrm{M}} \approx \mathbf{H}_{\mathrm{M}}\mathbf{U}
\end{array}
\end{equation}
where $\mathbf{U}_{\mathrm{H}}=\mathbf{UBS}$ and $\mathbf{H}_{\mathrm{M}}=\mathbf{RH}$. CNMF alternately unmixes $\mathbf{Y}_{\mathrm{H}}$ and $\mathbf{Y}_{\mathrm{M}}$ in the framework of nonnegative matrix factorization (NMF) \cite{LeeNature1999} to estimate $\mathbf{H}$ and $\mathbf{U}$ under the constraints of the relative sensor characteristics. CNMF starts with NMF unmixing of the low spatial resolution HS data. The matrix $\mathbf{H}$ can be initialized using, for example, the vertex component analysis (VCA) \cite{Nascimento2005}, and $\mathbf{H}$ and $\mathbf{U}_{\mathrm{H}}$ are then alternately optimized by minimizing $\| \mathbf{Y}_{\mathrm{H}}-\mathbf{H}\mathbf{U}_{\mathrm{H}} \|_F^2$ using Lee and Seung's multiplicative update rules \cite{LeeNature1999}. Next, $\mathbf{U}$ is estimated from the higher spatial resolution data. $\mathbf{H}_{\mathrm{M}}$ is set to $\mathbf{R}\mathbf{H}$ and $\mathbf{U}$ is initialized by the spatially up-sampled matrix of $\mathbf{U}_{\mathrm{H}}$ obtained by using bilinear interpolation. For HS pansharpening ($n_\lambda$=1), only $\mathbf{U}$ is optimized by minimizing $\| \mathbf{Y}_{\mathrm{M}}-\mathbf{H}_{\mathrm{M}}\mathbf{U} \|_F^2$ with the multiplicative update rule, whereas both $\mathbf{H}_{\mathrm{M}}$ and $\mathbf{U}$ are alternately optimized in the case of HS+MS data fusion. Finally, the high spatial resolution HS data can be obtained by the multiplication of $\mathbf{H}$ and $\mathbf{U}$. The abundance sum-to-one constraint is implemented using a method given in \cite{HeinzTGRS2001}, where the data and signature matrices are augmented by a row of constants. The relative sensor characteristics, such as $\mathbf{BS}$ and $\mathbf{R}$, can be estimated from the observed data sources \cite{YokoyaJSTARS2013}.

\section{Quality Assessment of fusion products}
Quality assessment of a pansharpened real-life HS image is not an easy task \cite{Wald97}, \cite{Chav91}, since a reference image is generally not available. When such an image is not available, two kinds of comparisons can be performed:
i) Each band of the fused image can be compared with the PAN image, with an appropriate criterion. The PAN image can also be compared with the PAN image reconstructed from the fused image. ii) The fused image can be spatially degraded to the resolution of the original HS image. The two images can then be compared, to assess to what extent the spectral information has been modified by the fusion method.

In order to be able to use a reference image for quality assessment, one normally has to resort to the use of semi-synthetic HS and PAN images. In this case, a real-life HS image is used as reference. The HS and PAN images to be processed are obtained by degrading this reference image. A common methodology for obtaining the degraded images is Wald's protocol, described in the next subsection. In order to evaluate the quality of the fused image with respect to the reference image, a number of statistical measures can be computed. The most widely used ones are described ahead, and used in the experiments reported in Section \ref{sec:results}.

\subsection{Wald's Protocol}
A general paradigm for quality assessment of fused images that is usually accepted in the research community was first proposed by Wald et al. \cite{Wald97}. This paradigm is based on two properties that the fused data have to have, as much as possible, namely consistency and synthesis properties. The first property requires the reversibility of the pansharpening process: it states that the original HS image should be obtained by properly degrading the pansharpened image. The second property requires that the pansharpened image be as similar as possible to the image of the same scene that would be obtained, by the same sensor, at the higher resolution. This condition entails that both the features of each single band and the mutual relations among bands have to be preserved. 
However, the definition of an assessment method that fulfills these constraints is still an open issue \cite{Amro2011}, \cite{Du07}, and closely relates to the general discussion regarding image quality assessment \cite{Wang02} and image fusion \cite{Alpa08}, \cite{Piella03}.

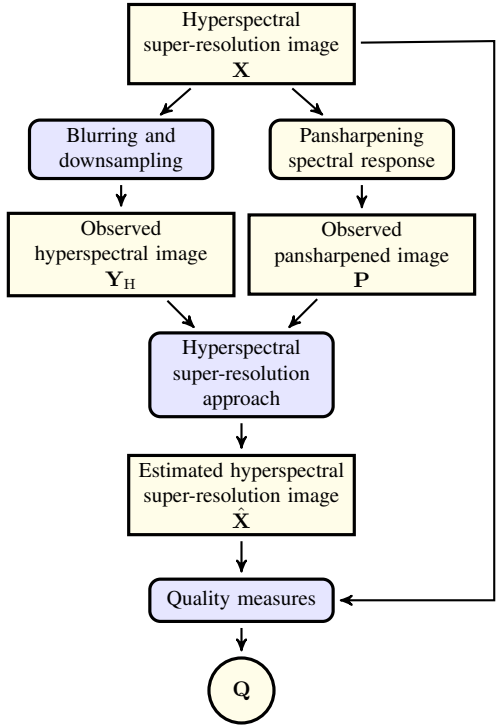
\begin{figure}
\begin{center}
\begin{tikzpicture}[scale=0.8,every node/.style={scale=0.8}]
\fill (2,6.5) node(HSR)[image] {Hyperspectral\\super-resolution image\\$\mathbf{X}$};
\fill (0,4.75) node(blurring)[process] {Blurring and downsampling};
\fill (4,4.75) node(specresp)[input] {Pansharpening\\ spectral response};
\fill (0,3) node(hyper)[image] {Observed\\hyperspectral image\\$\mathbf{Y}_{\mathrm{H}}$};
\fill (4,3) node(multi)[image] {Observed\\pansharpened image\\$\mathbf{P}$};
\fill (2,1) node(approach)[process] {Hyperspectral super-resolution approach};
\fill (2,-1) node(estHSR)[image] {Estimated hyperspectral\\super-resolution image\\$\hat{\mathbf{X}}$};
\fill (2,-2.75) node(quality)[process] {Quality measures};
\fill (2,-4.25) node(Q)[object] {$\mathbf{Q}$};
\draw [pil] (HSR) -- (blurring) {};
\draw [pil] (HSR) -- (specresp) {};
\draw [pil] (blurring) -- (hyper) {};
\draw [pil] (specresp) -- (multi) {};
\draw [pil] (hyper) -- (approach) {};
\draw [pil] (approach) -- (estHSR) {};
\draw [pil] (multi) -- (approach) {};
\draw [pil] (estHSR) -- (quality) {};
\draw [pil] (HSR) -- (6.2,6.55) -- (6.2,-2.75) -- (quality);
\draw [pil] (quality) -- (Q) {};
\end{tikzpicture}
\caption{\label{fig:wald_methodology}Flow diagram of the experimental methodology, derived from Wald's protocol (simulated observations), for synthetic and semi-real datasets.}
\end{center}
\end{figure}

Wald's protocol for assessing the quality of pansharpening methods~\cite{Wald97}, depicted in Fig.~\ref{fig:wald_methodology}, synthetically generates simulated observed images from a reference HS image, and then evaluates the pansharpening methods' results against that reference image. The protocol consists of the following steps:

\begin{itemize}

\item Given a HS image, $\mathbf{X}$, to be used as reference, a simulated observed low spatial resolution HS image, $\mathbf{Y}_{\mathrm{H}}$, is obtained by applying a Gaussian blur to $\mathbf{X}$, and then downsampling the result by selecting one out of every $d$ pixels in both the horizontal and vertical directions, where $d$ denotes the downsampling factor.

\item A simulated PAN image, $\mathbf{P}$, is obtained by multiplying the reference HS image, on the left, by a suitably chosen spectral response vector, $\mathbf P = \mathbf r^T \mathbf X$.

\item The pansharpening method to be evaluated is applied to the simulated observations $\mathbf{Y}_{\mathrm{H}}$ and $\mathbf{P}$, yielding the estimated superresolution HS image, $\hat{\mathbf{X}}$.

\item Finally, the estimated superresolution HS image and the reference one are compared, to obtain quantitative quality measures.

\end{itemize}

\subsection{Quality Measures}
\label{QualityMeasures}
Several quality measures have been defined in the literature, in order to determine the similarity between estimated and reference spectral images. These measures can be generally classified into three categories, depending on whether they attempt to measure the spatial, spectral or global quality of the estimated image. This review is limited to the most widely used quality measures, namely the \textit{cross correlation {\em (CC)}}, which is a spatial measure, the \textit{spectral angle mapper {\em (SAM)}}, which is a spectral measure, and the \textit{root mean squared error} (RMSE) and \textit{erreur relative globale adimensionnelle de synth\`ese} (ERGAS) \cite{ERGAS}, which are global measures. Below we provide the formal definitions of these measures operating on the estimated  image $\widehat{\bf X} \in \mathbb{R}^{m_\lambda \times n}$ and on the reference HS image $\MATima \in \mathbb{R}^{m_\lambda \times n}$. In the definitions, $\widehat{\bf x}_j$ and ${\bf x}_j$ denote the $j$th columns of $\widehat{\bf X}$ and ${\bf X}$, respectively, the matrices ${\bf A,B}\in\mathbb{R}^{1\times n}$ denote two generic single-band images, and ${\bf A}_i$ denotes the $i$th element of $\bf A$.

\subsubsection{Cross correlation} CC,  which characterizes  the geometric distortion, is defined as
\begin{align}
    \text{CC}(\widehat{\bf X},{\bf X}) & = \frac{1}{m_\lambda}\sum_{i=1}^{m_\lambda} \text{CCS}(\widehat{\bf X}^i,{\bf X}^i),
\end{align}
where CCS is the cross correlation for a single-band image, defined as
\begin{align*}
    \text{CCS}({\bf A, B}) & = \frac{\sum_{j=1}^n({\bf A}_j - \mu_{A})({\bf B}_j- \mu_{B})}
     {\sqrt{\sum_{j=1}^n({\bf A}_j - \mu_{A})^{2} \sum_{j=1}^n({\bf B}_j - \mu_{B})^{2}} },
\end{align*}
where, $\mu_{A} = (1/n)\sum_{j=1}^n{\bf A}_j$ is the sample mean of  ${\bf A}$.
The ideal value of CC is $1$.

\subsubsection{Spectral angle mapper} SAM, which is a spectral measure, is defined as
\begin{align}
    \text{SAM}(\widehat{\bf X},{\bf X}) & = \frac{1}{n}\sum_{j=1}^n \text{SAM}(\widehat{\bf x}_j,{\bf x}_j),
\end{align}
where, given the vectors ${\bf a,b}\in\mathbb{R}^{m_\lambda}$,
\begin{equation}
\textrm{SAM}({\bf a,b}) = \arccos\left(\dfrac{\langle {\bf a,b}\rangle}{\Vert {\bf a} \Vert \Vert {\bf b} \Vert}\right),
\end{equation}
$\langle {\bf a,b}\rangle = {\bf a}^T{\bf b}$ is inner product between $\bf a$ and $\bf b$, and $\Vert\cdot \Vert$ is the $\ell_{2}$ norm.
SAM is a measure of the spectral shape preservation. The optimal value of SAM is $0$. The values of SAM reported in our experiments have been obtained by averaging the values obtained for all the image pixels.

\subsubsection{Root mean squared error} The RMSE measures the $\ell_2$ error between the two matrices ${\bf X}$ and $\widehat{\bf X}$
\begin{equation}
\textrm{RMSE}(\widehat{\bf X},{\bf X}) = \dfrac{\| \widehat{\bf X}-{\bf X}\|_F}{\sqrt{n*m_{\lambda}}}\,
\end{equation}
where $\|{\bf X}\|_F=\sqrt{\text{trace}({\bf X}^T{\bf X})}$ is the Frobenius norm of $\bf X$. The ideal value of RMSE is $0$.

\subsubsection{Erreur relative globale adimensionnelle de synth\`ese} ERGAS offers a global indication of the quality of a fused image. It is defined as
\begin{equation}
\text{ERGAS}(\widehat{\bf X},{\bf X}) =
  100\,d\sqrt{\frac{1}{m_\lambda}\sum_{k=1}^{m_\lambda} \left(\frac{\text{RMSE}_k}{\mu_k} \right)^2},
\end{equation}
where $d$ is the ratio between the linear resolutions of the PAN and HS images, defined as
$$
 d = \frac{\text{PAN linear spatial resolution}}{\text{HS linear spatial resolution}},
$$
$\text{RMSE}_k = \dfrac{\| \widehat{\bf X}^k-{\bf X}^k\|_F}{\sqrt{n}}$,  $\mu_k$ is the sample mean of the $k$th band of ${\bf X}$. The ideal value of ERGAS is $0$.

\section{Experimental Results}
\subsection{Datasets}
\label{dataset}
The datasets that were used in the experimental tests were all semi-synthetic, generated according to the Wald's protocol. In all cases, the spectral bands corresponding to the absorption band of water vapor, and the bands that were too noisy, were removed from the reference image before further processing. Three real-life HS images have been used as reference images for the Wald protocol. In the following, we describe the datasets that were generated from these images. Table \ref{table_dataset} summarizes their properties. These datasets are expressed in spectral luminance (nearest to the sensor output, without pre-processing) and are correctly registered.

\begin{table}[!h]
\renewcommand{\arraystretch}{1.3}
\caption{Caracteristic of the three datasets}
\label{table_dataset}
\centering
\small
\begin{tabular}{|c|c|c|c|c|c|c|}
\hline
\textbf{dataset} & \textbf{dimensions} & \textbf{spatial res} & \textbf{N}& \textbf{instrument}\\
\hline
Moffett & \begin{tabular}[c]{@{}l@{}}PAN $185\times395$\\HS $37\times79$\end{tabular} & \begin{tabular}[c]{@{}l@{}}$20$m\\$100$m\end{tabular} & $224$ & AVIRIS \\
\hline
Camargue & \begin{tabular}[c]{@{}l@{}}PAN $500\times500$\\HS $100\times100$\end{tabular} & \begin{tabular}[c]{@{}l@{}}$4$m\\$20$m\end{tabular}  & $125$  & HyMap\\
\hline
Garons & \begin{tabular}[c]{@{}l@{}}PAN $400\times400$\\HS $80\times80$\end{tabular} & \begin{tabular}[c]{@{}l@{}}$4$m\\$20$m\end{tabular}  & $125$ & HyMap \\
\hline
\end{tabular}
\normalsize
\end{table}

\subsubsection{Moffett field dataset}
This dataset represents a mixed urban/rural scene. The dimensions of the PAN are $185\times395$ with a spatial resolution of $20$m whereas the size of the HS image is $37\times79$ with a spatial resolution of $100$m (which means a spatial resolution ratio of $5$ between the two images). The HS image has been acquired by the airborne hyperspectral instrument airborne visible infrared image spectrometer (AVIRIS). This instrument is characterized by $224$ bands covering the spectral range $0.4-2.5\mu$m.

\subsubsection{Camargue dataset}
This dataset represents a rural area with different kind of crops. The dimensions of the PAN image are $500\times500$ pixels with a spatial resolution of $4$m whereas the size of the HS image is $100\times100$ pixels with a spatial resolution of $20$m, (which means a spatial resolution ratio of $5$ between the two images). The HS image has been acquired by the airborne hyperspectral instrument HyMap (Hyperspectral Mapper) in 2007. The hyperspectral instrument is characterized by $125$ bands covering the spectral range $0.4-2.5\mu$m.

\subsubsection{Garons dataset}
This dataset represents a rural area with a small village. The dimension of the PAN image are $400\times400$ with a spatial resolution of $4$m whereas the size of the HS image is $80\times80$ with a spatial resolution of $20$m, (which means a spatial resolution ratio of $5$ between the two images). This dataset has been acquired with the HyMap instrument in 2009.

\subsection{Results and Discussion}
\label{sec:results}

Methods presented in Section \ref{sec:HS_pan:tech} have been applied on the three datasets presented in Section \ref{dataset} and analyzed following the Wald's Protocol (Section \ref{QualityMeasures}). Tables \ref{table_dataset1}, \ref{table_dataset2}, \ref{table_dataset3} report their quantitative evaluations with respect to the quality measures detailed in Section \ref{QualityMeasures}.

\begin{table}[h!]
\renewcommand{\arraystretch}{1.3}

\caption{Quality measures for the Moffett field dataset}
\label{table_dataset1}
\centering
\small
\begin{tabular}{|c|c|c|c|c|}
\hline
\textbf{method} & \textbf{CC} & \textbf{SAM} & \textbf{RMSE}& \textbf{ERGAS}\\
\hline
SFIM & 0.92955 & 9.5271 & 365.2577 & 6.5429\\
\hline
MTF-GLP & 0.93919 & 9.4599 & 352.1290 & 6.0491\\
\hline
MTF-GLP-HPM & 0.93817 & 9.3567 & 354.8167 & 6.1992\\
\hline
GS & 0.90521 & 14.1636 & 443.4351 & 7.5952\\
\hline
GSA & 0.93857 & 11.2758 & 363.7090 & 6.2359\\
\hline
PCA & 0.89580 & 14.6132 & 463.2204 & 7.9283\\
\hline
GFPCA & 0.91614 & 11.3363 & 404.2979 & 7.0619\\
\hline
CNMF & 0.95496 & 9.4177 & 314.4632 & 5.4200\\
\hline
Bayesian Naive & 0.97785 & 7.1308 & 220.0310 & 3.7807\\
\hline
Bayesian Sparse & \textbf{0.98168} & \textbf{6.6392} & \textbf{200.3365} & \textbf{3.4281} \\
\hline
HySure & 0.97059 & 7.6351 & 254.2005  & 4.3582 \\
\hline
\end{tabular}
\normalsize
\end{table}

\begin{table}[h!]
\renewcommand{\arraystretch}{1.3}
\caption{Quality measures for the Camargue dataset}
\label{table_dataset2}
\centering
\small
\begin{tabular}{|c|c|c|c|c|}
\hline
\textbf{method} & \textbf{CC} & \textbf{SAM} & \textbf{RMSE}& \textbf{ERGAS}\\
\hline
SFIM & 0.91886 & 4.2895 & 637.1451 & 3.4159\\
\hline
MTF-GLP & 0.92397 & 4.3378 & 622.4711 & 3.2666\\
\hline
MTF-GLP-HPM & 0.92599 & 4.2821 & 611.9161 & 3.2497\\
\hline
GS & 0.91262 & 4.4982 & 665.0173 & 3.5490\\
\hline
GSA & 0.92826 & 4.1950 & 587.1322 & 3.1940\\
\hline
PCA & 0.90350 & 5.1637 & 710.3275 & 3.8680\\
\hline
GFPCA & 0.89042 & 4.8472 & 745.6006 & 4.0001\\
\hline
CNMF & 0.9300 & 4.4187 & 591.3190 & 3.1762\\
\hline
Bayesian Naive & 0.95195 & 3.6428 & 489.5634 & 2.6286\\
\hline
Bayesian Sparse & \textbf{0.95882} & \textbf{3.3345} & \textbf{448.1721} & \textbf{2.4712}\\
\hline
HySure & 0.9465 & 3.8767 & 511.8525 & 2.8181\\
\hline
\end{tabular}
\normalsize
\end{table}

\begin{table}[h!]
\renewcommand{\arraystretch}{1.3}
\caption{Quality measures for the Garons dataset}
\label{table_dataset3}
\centering
\small
\begin{tabular}{|c|c|c|c|c|}
\hline
\textbf{method} & \textbf{CC} & \textbf{SAM} & \textbf{RMSE}& \textbf{ERGAS}\\
\hline
SFIM & 0.77052 & 6.7356 & 1036.4695 & 5.1702\\
\hline
MTF-GLP & 0.80124 & 6.6155 & 956.3047 & 4.8245\\
\hline
MTF-GLP-HPM & 0.79989 & 6.6905 & 962.1076 & 4.8280\\
\hline
GS & 0.80347 & 6.6627 & 1037.6446 & 5.1373\\
\hline
GSA & 0.80717 & 6.7719 & 928.6229 & 4.7076\\
\hline
PCA & 0.81452 & 6.6343 & 1021.8547 & 5.0166\\
\hline
GFPCA & 0.63390 & 7.4415 & 1312.0373 & 6.3416\\
\hline
CNMF & 0.82993 & 6.9522 & 893.9194 & 4.4927\\
\hline
Bayesian Naive & 0.86857 & 5.8749 & 784.1298 & 3.9147\\
\hline
Bayesian Sparse & \textbf{0.87834} & \textbf{5.6377} & \textbf{750.3510} & \textbf{3.7629}\\
\hline
HySure & 0.86080 & 6.0224 & 778.1051 & 4.0454\\
\hline
\end{tabular}
\normalsize
\end{table}

\begin{figure}[!h]
\includegraphics[width=1\linewidth]{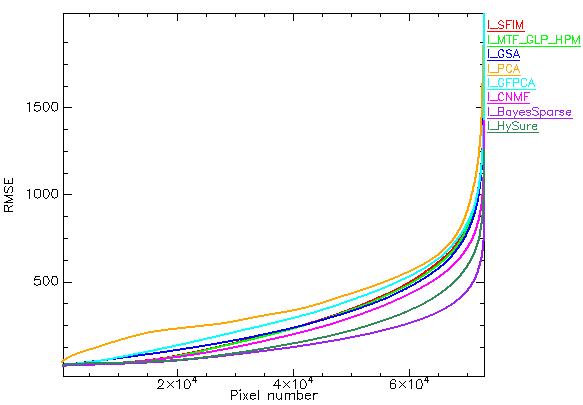}
\caption{RMSE between the methods' result and the reference image, per pixel for the Moffett field dataset}
\label{RMSE1}
\end{figure}

\begin{figure}[!h]
\includegraphics[width=1\linewidth]{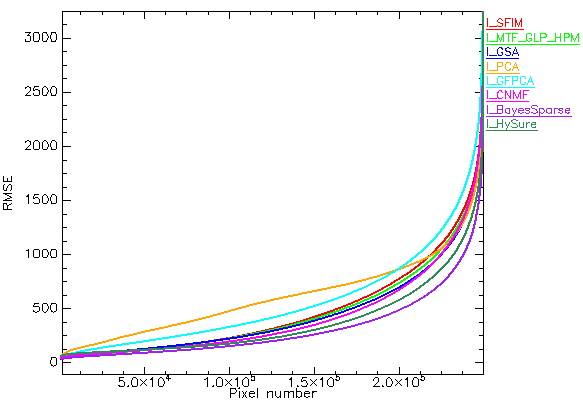}
\caption{RMSE between the methods' result and the reference image, per pixel for the Camargue dataset}
\label{RMSE2}
\end{figure}

\begin{figure}[!h]
\includegraphics[width=1\linewidth]{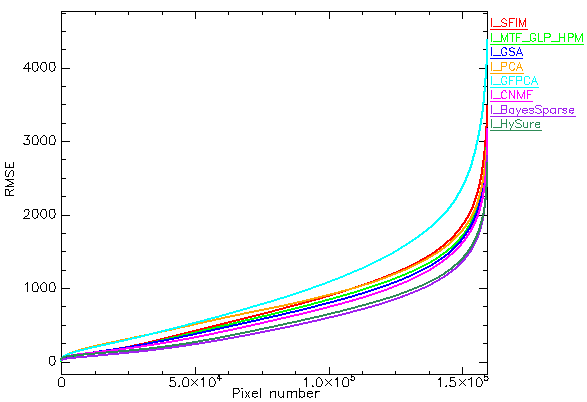}
\caption{RMSE between the methods' result and the reference image, per pixel for the Garons dataset}
\label{RMSE3}
\end{figure}

Figures \ref{RMSE1}, \ref{RMSE2}, \ref{RMSE3} represent the RMSEs per pixel between the image estimated by some methods and the reference image for the three considered datasets. Note that, for sake of conciseness, some methods have not been considered here but only their improved versions are presented. More specifically, GS has been removed since GSA is an improved version of GS. Indeed, GSA is expected to give better results than GS thanks to its adaptive estimation of the weight for generating the equivalent PAN image from the HS image, which allows the spectral distortion to be reduced. Bayesian naive approach has been also removed since the sparsity-based approach relies on a more complex prior and gives also better results. MTF-GLP and MTF-GLP-HPM yield similar results so only the latter has been considered.

\begin{figure}[!h]
\includegraphics[width=1\linewidth]{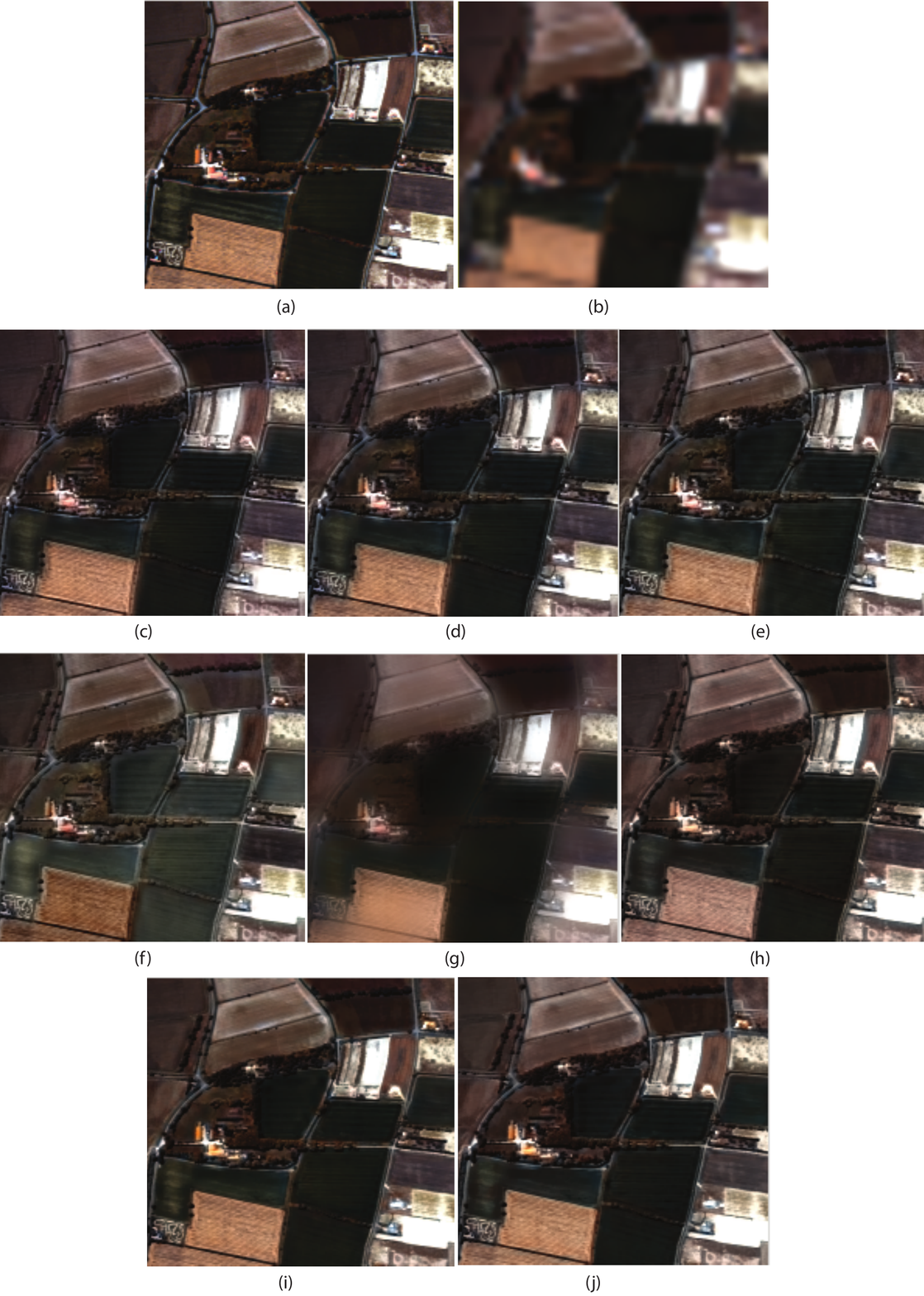}
\caption{Details of original and fused Camargue dataset HS image in the visible domain. (a) reference image, (b) interpolated HS image, (c) SFIM, (d) MTF-GLP-HPM, (e) GSA, (f) PCA, (g) GFPCA, (h) CNMF, (i) Bayesian Sparse, (j) HySure}
\label{result}
\end{figure}

\begin{figure}[!h]
\includegraphics[width=1\linewidth]{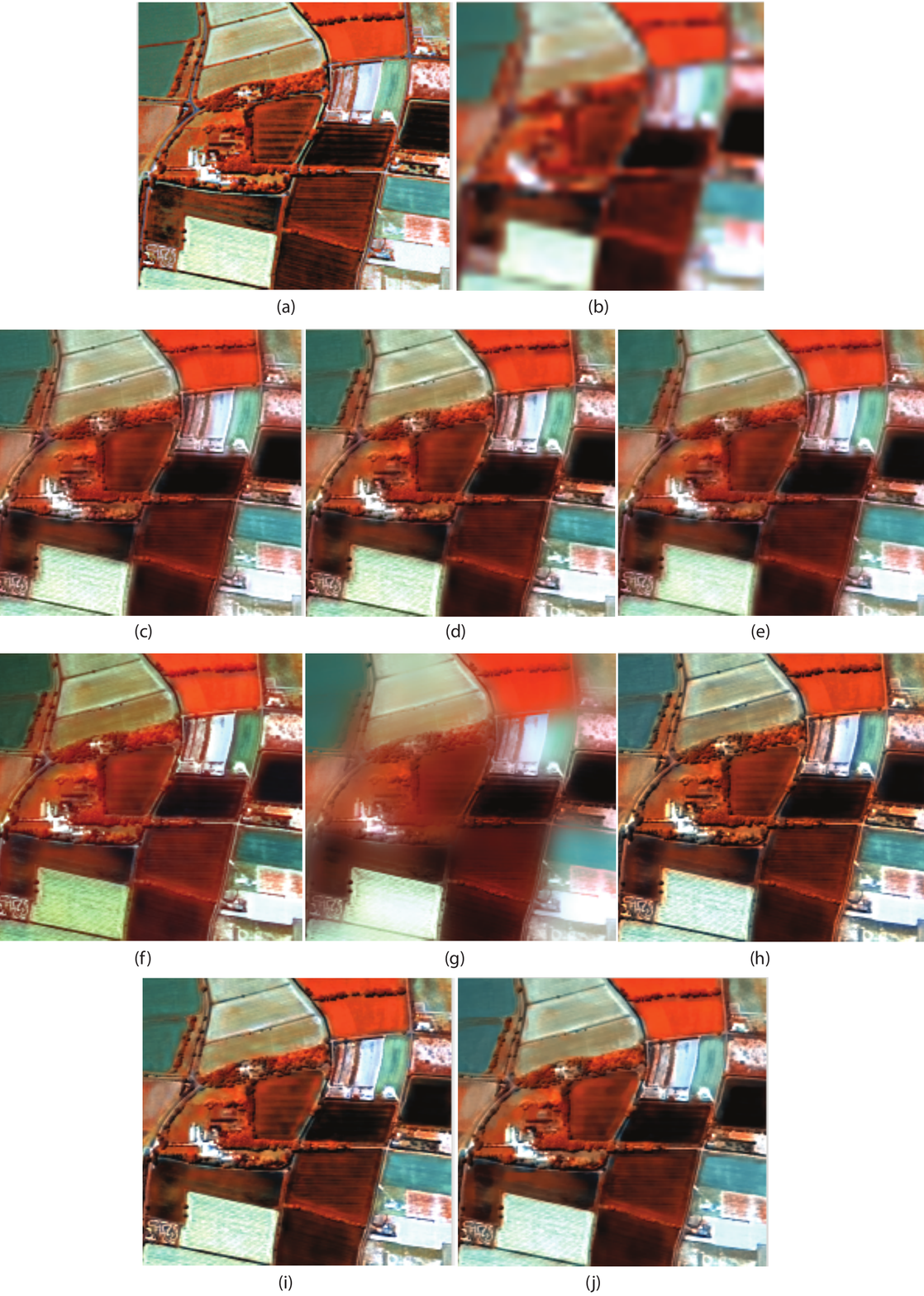}
\caption{Details of original and fused Camargue dataset HS image in the SWIR domain. (a) reference image, (b) interpolated HS image, (c) SFIM, (d) MTF-GLP-HPM, (e) GSA, (f) PCA, (g) GFPCA, (h) CNMF, (i) Bayesian Sparse, (j) HySure}
\label{result_swir}
\end{figure}

Figures \ref{result} and \ref{result_swir} show extracts of the final result obtained by the considered methods on the Camargue dataset in the visible (R$=704.39$nm, G$=557.90$nm, B$=454.5$nm) and in the SWIR (R$= 1216.7$nm, G$=1703.2$nm, B$=2159.8$nm) domains, respectively.

\begin{figure}[!h]
\includegraphics[width=1\linewidth]{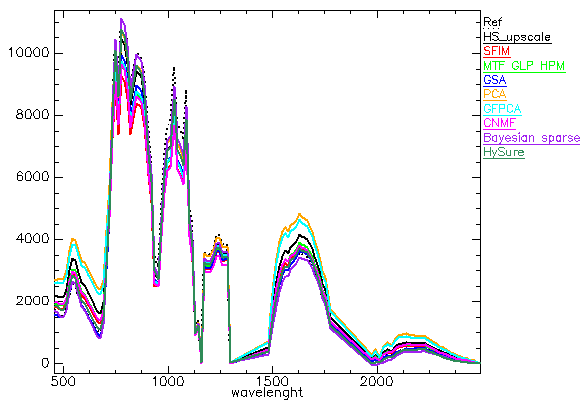}
\caption{Luminance of the pixel corresponding to the 10th percentile of the RMSE (Camargue dataset)}
\label{RMSE10}
\end{figure}

\begin{figure}[!h]
\includegraphics[width=1\linewidth]{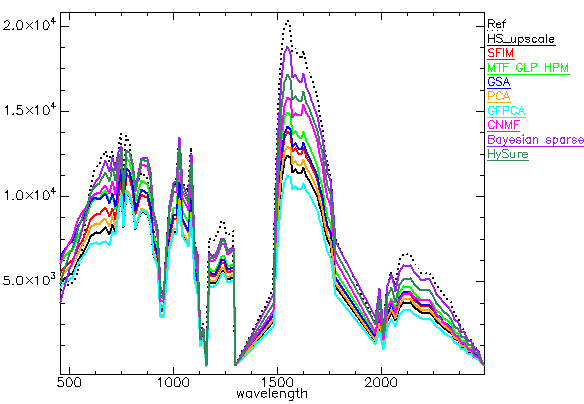}
\caption{Luminance of the pixel corresponding to the 50th percentile of the RMSE (Camargue dataset)}
\label{RMSE50}
\end{figure}

\begin{figure}[!h]
\includegraphics[width=1\linewidth]{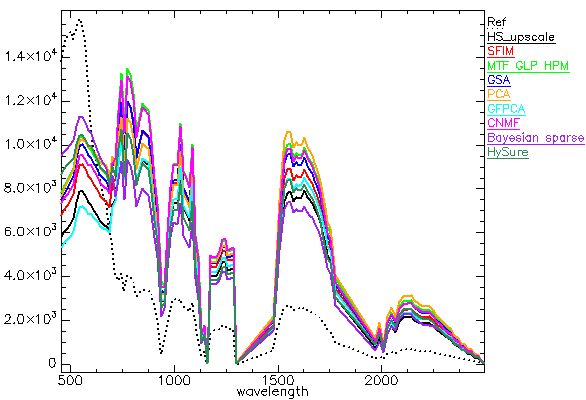}
\caption{Luminance of the pixel corresponding to the 90th percentile of the RMSE (Camargue dataset)}
\label{RMSE90}
\end{figure}

Figures \ref{RMSE10}, \ref{RMSE50} and \ref{RMSE90} show pixel spectra recovered by the fusion methods, which correspond to $10$th, $50$th and $90$th percentile of RMSE, respectively. Those spectra have been selected by choosing GSA as the reference for RMSE value. GSA have been chosen since it is a classical approach that has been widely used in literature and also gives good results. To ensure a reasonable number of figures, only visual results and some spectra of the Camargue dataset has been reported in this article. The results for the two other datasets can be found in the supporting document \cite{Loncan2015supporting} available online\footnote{\url{http://openremotesensing.net}}. In particular, because of the nature of the Garons dataset (village with lot of small buildings) and the chosen ratio of $5$, worse results have been obtained than for the two first datasets since a lot of mixing is presented in the HS image.

A visual analysis of the result shows that most of the fusion approaches considered in this paper give good results, excepted two methods: PCA and GFPCA. PCA belongs to the class of CS methods which are known to be characterized by their high fidelity in rendering the spatial details but their generation of significant spectral distortion. This is clearly visible in Figure \ref{result} (f), where significant differences of color can be observed with respect to the reference image, in particular when examining the different fields. GFPCA here also performs poorly. Compared with PCA, there is less spectral distortion but the included spatial information seems to be not sufficient, since the fused image is significantly blurred. Spatial information provided by PCA is better since the main information of HS image (where the spatial information is contained) is replaced by the high spatial information contained in the PAN image. When using GFPCA, the guided filter controls the amount of spatial information added to the data, so not all the spatial information may be added to avoid to modify the spectral information too much. For the Moffett field dataset, GFPCA performs a little bit better since, in this dataset, there is a lot of large areas. Thus blur is less present whereas, in the Garons dataset, GFPCA performs worse since this image consists of numerous small features, leading to more blurring effects. As a consequence, in this case, GFPCA performs worse than PCA.

To analyze the spectrum in detail, chosen thanks to RMSE percentiles, some additional information about the corresponding pixels are needed. Fig. \ref{RMSE50} corresponds to a pixel in the reference image which represents a red building. Since in the HS image this building is mixed with its neighborhood, we do not have the same information between the reference image (``pure'' spectrum) and the HS image (``mixed" spectrum). Fig. \ref{RMSE10} corresponds to a pixel in a vegetation area, no mixing is present and very good results have been obtained for all the methods. For Fig. \ref{RMSE90}, the pixel belongs to a small building not visible in the HS image and spectral mixing is then also present. More generally, spectra in the HS and reference images differ since some mixing processes occur in the HS image. Thus, the HS pansharpening methods are expected to provide spectra that are more similar to the HS spectra (which contains the available spectral information) than the reference (which has information missing in the HS which should not be found in the result, unless successful unmixing has been conducted). However, it is important to note that for Fig.\ref{RMSE50}, Bayesian methods and HySure successfully recover spectra that are more similar to the reference spectrum than the HS spectrum.

Table \ref{table_computational time} report the computational times required by each HS pansharpening methods on the Camargue dataset those values have been obtained with a Intel Core i5 3230M 2.6 GHz with 8 GB RAM. Based on this table, these methods can be classified as follows:
\begin{itemize}
\item Methods which do not work well for HS pansharpening: PCA, GS, GFPCA
\item Methods which work well with a low time of computation (few seconds): GSA, MRA methods, Bayesian Naive
\item Methods which work well with an average time of computation (around one minute): CNMF
\item Methods which work well (slightly better) with an important time of computation (few minutes, depends greatly on the size of the dataset): Bayesian Sparse and HySure.
\end{itemize}

\begin{table}[!h]
\renewcommand{\arraystretch}{1.3}
\caption{Computational times of the different methods (in seconds)}
\label{table_computational time}
\centering
\small
\begin{tabular}{|c|c|c|c|}
\hline
\textbf{method} & \textbf{Moffett} & \textbf{Camargue} & \textbf{Garons}\\
\hline
SFIM & 1.26 & 3.47 & 2.74\\
\hline
MTF-GLP & 1.86 & 4.26 & 4.00\\
\hline
MTF-GLP-HPM & 1.71 & 4.25 & 2.98\\
\hline
GS & 4.77 & 8.29 & 5.56\\
\hline
GSA & 5.52 & 8.73 & 5.99\\
\hline
PCA & 3.46 & 8.92 & 6.09\\
\hline
GFPCA & 2.58 & 8.51 & 4.36\\
\hline
CNMF & 10.98 & 47.54 & 23.98\\
\hline
Bayesian Naive & 1.31 & 7.35 & 3.07\\
\hline
Bayesian Sparse & 133.61 & 485.13 & 259.44\\
\hline
HySure & 140.05 & 296.27 & 177.60\\
\hline
\end{tabular}
\normalsize
\end{table}

To summarize, the comparison of the different methods performances for RMSE curves and quality measures confirms than PCA and GFPCA does not provide good results for HS pansharpening (GFPCA is know to perform much better on HS+RGB data). The other methods perform well, with Bayesian approaches having slightly better result but with a higher computational cost. The favorable fusion performance obtained by the Bayesian methods can be explained, in part, by the fact that they rely on a forward modeling of the PAN and HS images and explicitly exploit the spatial and spectral degradations applied to the target image. However, these algorithms may suffer from performance discrepancies when the parameters of these degradations (i.e., spatial blurring kernel, sensor spectral response) are not perfectly known. In particular, when these parameters are fully unknown and need to be fixed, they can be estimated jointly with the fused image, as in \cite{Wei2014ICIP}, or estimated from the MS and HS images in a pre-processing step, following the strategies in \cite{YokoyaJSTARS2013} or \cite{Simoes2014b}. CS methods are fast to compute and easy to implement. They provide good spatial results but poor spectral results with significant spectral distortions, in particular when considering PCA and GS. GSA provides better results than the two other methods thanks to its adaptive weight estimation reducing the spectral distortion of the equivalent PAN image created from the HS image. MRA methods are fast, MTF-based methods give better results than SFIM and perform as well as the most competitive algorithms with higher computational complexity. SFIM does not perform as well than the other MRA methods since it used a box filter which should give less good result. In our experimentations, results from SFIM are not so different from those obtained with the MTF-based methods. This may come from the fact that semi-synthetic datasets are used so MTF may not be fully used to its potential. Table \ref{pro_cons} report these pro and cons associated with each HS pansharpening method.

\begin{table}[!h]
\setlength{\tabcolsep}{0.3mm}
\renewcommand{\arraystretch}{1.2}
\caption{Pros and Cons of each methods}
\label{pro_cons}
\centering
\small
\begin{tabular}{|c|l|l|}
\hline
\textbf{method} & \textbf{pros} & \textbf{cons} \\
\hline
\begin{tabular}[c]{@{}c@{}} SFIM \\ II.B.1 \end{tabular} & \begin{tabular}[c]{@{}l@{}}1) Low computational\\ complexity\end{tabular} & \begin{tabular}[c]{@{}l@{}}1) Reduced performance \\when compared to MTF \\ methods (since it uses a\\ box filter)\end{tabular}\\
\hline
\begin{tabular}[c]{@{}c@{}} MTF-GLP \\II.B.2 \end{tabular}& \begin{tabular}[c]{@{}l@{}}1) Performs well\\2) Low computational\\ complexity \end{tabular} &
 \\
\hline
\begin{tabular}[c]{@{}c@{}} MTF-GLP-HPM\\ II.B.2 \end{tabular}& \begin{tabular}[c]{@{}l@{}}1) Performs well\\2) Low computational\\ complexity\end{tabular} &
 \\
\hline
\begin{tabular}[c]{@{}c@{}} GS \\ II.A.2 \end{tabular}& \begin{tabular}[c]{@{}l@{}}1) Spatial information \\ is well preserved\\2) Low computational\\ complexity \\ 3) Easy implementation \end{tabular}
 & \begin{tabular}[c]{@{}l@{}} 1) Low performance\\ for HS images \\ 2) Significant spectral\\ distortion \end{tabular}
\\
\hline
\begin{tabular}[c]{@{}c@{}} GSA \\ II.A.2 \end{tabular}& \begin{tabular}[c]{@{}l@{}}1) Spatial information\\is well preserved\\2) Spectral distortion\\ is reduced (compared\\ to GS)\\ 3) Low computational\\ complexity\\ 4) Easy implementation \end{tabular} & \\
\hline
\begin{tabular}[c]{@{}c@{}} PCA \\ II.A.1 \end{tabular}& \begin{tabular}[c]{@{}l@{}}1) Spatial information \\ is well preserved\\2)Low computational\\ complexity \\ 3) Easy implementation \end{tabular}
 & \begin{tabular}[c]{@{}l@{}} 1) Low performance \\for HS images \\ 2) Significant spectral\\ distortion \end{tabular}
\\
\hline
\begin{tabular}[c]{@{}c@{}} GFPCA \\ II.C.1 \end{tabular}&  \begin{tabular}[c]{@{}l@{}}1) Spectral information\\ is well preserved \\ 2) Low computational\\ complexity \end{tabular}
 &  \begin{tabular}[c]{@{}l@{}}1) Low performance\\for HS images (work \\ better with RGB images)\\ 2) Not enough spatial\\ information added (lot \\of blur) \end{tabular}
\\
\hline
\begin{tabular}[c]{@{}c@{}} CNMF \\ II.E.1 \end{tabular}& \begin{tabular}[c]{@{}l@{}} 1) Good results \\(spatial and spectral)\end{tabular} & \begin{tabular}[c]{@{}l@{}}1) Sensor characteristics\\ required \\2) Medium computational\\ cost\end{tabular}\\
\hline
\begin{tabular}[c]{@{}c@{}} Bayesian Naive\\ II.D.1 \end{tabular}&\begin{tabular}[c]{@{}l@{}} 1) Good results \\(spatial and spectral)\\ 2)  Low computational \\ complexity \end{tabular} & \begin{tabular}[c]{@{}l@{}}1) Sensor characteristics\\ required \end{tabular}  \\
\hline
\begin{tabular}[c]{@{}c@{}} Bayesian Sparse \\ II.D.2 \end{tabular}& \begin{tabular}[c]{@{}l@{}} 1) Good results \\(spatial and spectral)\end{tabular} & \begin{tabular}[c]{@{}l@{}}1) high computational\\ cost \\ 2) Sensor characteristics\\ required \end{tabular}\\
\hline
\begin{tabular}[c]{@{}c@{}} HySure \\ II.D.3 \end{tabular}& \begin{tabular}[c]{@{}l@{}} 1) Good results \\(spatial and spectral)\end{tabular} &\begin{tabular}[c]{@{}l@{}} 1) high computational\\ cost\end{tabular}\\
\hline
\end{tabular}
\normalsize
\end{table}

Finally, note that, in our experimentations, no registration error and temporal misalignment have been considered, which suggests that the robustness of the different methods has not been fully analyzed. When such problems may occur, CS and MRA methods may perform better thanks to their great robustness. In particular, CS methods are robust against misregistration error and aliasing, whereas MRA approaches are robust against aliasing and temporal misalignment. It is also worthy to note that the quality of a fusion method should also be related to a specific application (such as classification or target detection). Indeed, a method providing images with good performance metrics may or may not be the best for this specific application.

\section{Conclusion}
In this paper a qualitative and quantitative comparison of 11 different HS pansharpening methods was conducted, considering classical MS pansharpening techniques adapted to the HS context, and methods originally designed for HS pansharpening. More precisely, five classes of methods were presented: CS, MRA, Hybrid, Bayesian and matrix factorization. Those methods were evaluated on three different datasets representative of various scenario: mixed urban/rural area, rural area and urban area.

A careful analysis of their performances suggested a classification of these methods into four groups: i) Methods with poor fusion results (CS-based methods and GFPCA), ii) Methods with good fusion performances and low computational costs (MRA methods, GSA and Bayesian naive) that may be suitable for fusing large scale images, which is often the case for spaceborne hyperspectral imaging missions, iii) Methods with good fusion performances and  reasonable computational costs (CNMF), iv) Methods with slightly better fusion results but with higher computational costs (HySure and Bayesian Sparse). Those results were obtained with semi-synthetic datasets with no registration error or temporal misalignment. Thus robustness of the methods against these issues were not taken into account. When such problems may happen, different results could be obtained and classical pansharpening methods (CS and MRA) may give better results thanks to their robustness to these specific issues.

The experiments and the quality measures presented in this paper were performed using MATLAB implementations of the algorithms. A MATLAB toolbox is made available online\footnote{\url{http://openremotesensing.net}} to the community to help improving and developing new HS pansharpening methods and to facilitate comparison of the different methods.

\section{Acknowledgment}
The Garons and Camargue datasets were acquired in the frame of the PRF Enviro program (internal federative project lead at ONERA). This work was partially supported by the Fundac{c}~{a}o para a Ci\^encia e Tecnologia, Portuguese Ministry of Science and Higher Education (UID/EEA/50008/2013), project PTDC/EEI-PRO/1470/2012, and grant SFRH/BD/87693/2012. Part of this work has been also supported by the Hypanema ANR Project n$^\circ$ANR-12-BS03-003 and by ANR-11-LABX-0040-CIMI within the program ANR-11-IDEX-0002-02. This work was supported by the SBO-IWT project Chameleon: Domain-specific Hyperspectral Imaging Systems for Relevant Industrial Applications, and FWO project G037115N ``Data fusion for image analysis in remote sensing''. This work is supported by China Scholarship Council. This work is supported by DGA (Direction Generale de l'Armement). This work was finally supported by the ERC CHESS (CHallenges in Extraction and Separation of Sources) and by ANR HYEP ANR 14-CE22-0016-01 (Agence National de la Recherche, Hyperspectral Imagery for Environmental Planning).

\ifCLASSOPTIONcaptionsoff
  \newpage
\fi

\bibliographystyle{ieeetran}

\bibliography{strings_all_ref,biblio_all_updated}

%
\end{document}

%% file: notations.tex
\input com_notations.tex

\newcommand{\MATima}{\bfX}

\newcommand{\nbbandima}{m_{\lambda}}

\newcommand{\imam}[1]{\boldsymbol{\mu}_{#1}}

\newcommand{\Covsub}{\boldsymbol{\Sigma}}

\newcommand{\imacovmat}[1]{\boldsymbol{\Sigma}_{#1}}

\newcommand{\Id}[1]{\textbf{I}_{#1}}

\newcounter{algo}
\renewcommand{\thealgo}{\arabic{algo}}

%% file: com_notations.tex
\def\bfA{{\mathbf{A}}}
\def\bfB{{\mathbf{B}}}

\def\bfD{{\mathbf{D}}}

\def\bfH{{\mathbf{H}}}

\def\bfN{{\mathbf{N}}}

\def\bfR{{\mathbf{R}}}
\def\bfS{{\mathbf{S}}}

\def\bfU{{\mathbf{U}}}

\def\bfX{{\mathbf{X}}}
\def\bfY{{\mathbf{Y}}}

\def\calP{{\mathcal{P}}}

\def\calN{{\mathcal{N}}}

\def\calP{{\mathcal{P}}}

\def\bsu{{\boldsymbol{u}}}

\def\bsx{{\boldsymbol{x}}}

\def\wtm{\widetilde{m}}